\documentclass[10pt,twocolumn,letterpaper]{article}

\usepackage{cvpr}
\usepackage{times}
\usepackage{epsfig}
\usepackage{graphicx}
\usepackage{amsmath}
\usepackage{amssymb}

\usepackage[numbers]{natbib}
\usepackage{subcaption}
\usepackage{xspace}
\usepackage{booktabs}
\usepackage{grffile}
\usepackage{authblk}

\newcommand{\onebyone}{$1\!\times\!1$\xspace}
\newcommand{\mobilenet}{MobileNetV2\xspace}
\newcommand{\resnext}{ResNeXt\xspace}

\usepackage[pagebackref=true,breaklinks=true,colorlinks,bookmarks=false]{hyperref}

\cvprfinalcopy 

\def\cvprPaperID{****} 

\begin{document}

\title{Dense Pruning of Pointwise Convolutions in the Frequency Domain}

\author{Mark Buckler}
\author{Neil Adit}
\author{Yuwei Hu}
\author{Zhiru Zhang}
\author{Adrian Sampson}

\affil{Computer Systems Lab, Cornell University, Ithaca, NY, USA \authorcr
   \tt markabuckler@gmail.com, \{\tt na469, yh457, zhiruz, asampson\}@cornell.edu}


\maketitle

\begin{abstract}
   Depthwise separable convolutions and frequency-domain convolutions are two
	recent ideas for building efficient convolutional neural networks. They are
	seemingly incompatible: the vast majority of operations in depthwise
	separable CNNs are in pointwise convolutional layers, but pointwise layers
	use \onebyone~ kernels, which do not benefit from frequency transformation.
	This paper unifies these two ideas by transforming the \emph{activations},
	not the kernels. Our key insights are that 1) pointwise convolutions commute
	with frequency transformation and thus can be computed in the frequency
	domain without modification, 2) each channel within a given layer has a
	different level of sensitivity to frequency domain pruning, and 3) each
	channel's sensitivity to frequency pruning is approximately monotonic with
	respect to frequency. We leverage this knowledge by proposing a new
	technique which wraps each pointwise layer in a discrete cosine transform
	(DCT) which is truncated to selectively prune coefficients above a given
	threshold as per the needs of each channel. To learn which frequencies
	should be pruned from which channels, we introduce a novel learned parameter
	which specifies each channel's pruning threshold. We add a new
	regularization term which incentivizes the model to decrease the number of
	retained frequencies while still maintaining task accuracy. Unlike weight
	pruning techniques which rely on sparse operators, our contiguous frequency
	band pruning results in fully dense computation. We apply our technique to
	\mobilenet and in the process reduce computation time by 22\% and incur
	\textless1\% accuracy degradation.
\end{abstract}

\section{Introduction}

Since the advent of convolutional neural networks for vision,
computational efficiency has been a primary concern. A broad swath of techniques
have successfully decreased the time and space required for CNN inference, such
as model compression and pruning~\cite{deepcompression,squeezenet},
frequency-domain computation~\cite{freq_domain_pruning}, and depthwise separable
convolutions~\cite{xception}. The resulting highly efficient networks are
particularly relevant to industrial deployments of vision, especially in
embedded and mobile settings where energy is a scarce resource.


High-efficiency CNNs have recently trended toward pointwise \onebyone{}
convolutions because of the high cost of larger kernels.
SqueezeNet~\cite{squeezenet}, replaced many 3$\times$3 kernels in a
traditional architecture with \onebyone{} kernels, and depthwise separable CNNs
replace traditional 3$\times$3 convolutions with a combination of pointwise convolutions and \emph{depthwise} 3$\times$3
convolutions. The recent proliferation of depthwise separable
CNNs~\cite{xception,mobilenetv1,mobilenetv2,shufflenet,shufflenetv2} has
demonstrated the effectiveness of this approach.

A concurrent but independent research direction has focused on \emph{weight
pruning} as a mechanism for compressing CNNs, based on absolute magnitude of parameters ~\cite{deepcompression} and frequency-domain based compression ~\cite{kilian-compress-cnns, freq_domain_pruning}.
However, weight pruning typically results in a \emph{sparse}
model which can be
counterintuitively slower than the dense computations found in unpruned models.
A recent study found, that even pruning 89\% of
AlexNet's~\cite{alexnet} weights resulted in a 25\%
\emph{slowdown}~\cite{yu2017scalpel}.


While depthwise separable networks and frequency-based weight pruning both
reduce parameter volume and computational cost, they are not trivially
compatible. The \onebyone{} filters that dominate computation in these
architectures are not amenable to a frequency-domain transform because they
contain no spatial context, and the 3$\times$3 filters account for an order of
magnitude less computation---so pruning them will have a negligible impact on
overall performance. The central goal in this paper is to properly combine these
two trends by using frequency-domain computation and pruning to speed up the
costly pointwise convolutions. In doing so, we make the following observations:
\begin{itemize}
	\item Pointwise convolutions, the primary computational bottleneck
		in modern efficient CNNs, can be computed in the frequency-domain
		\emph{without modification}.
	\item Certain frequency coefficients for certain channels can be pruned,
		resulting in fewer necessary operations.
	\item Any given channel's sensitivity to frequency pruning is
		approximately monotonic with respect to frequency.
\end{itemize}
These observations inspired our contributions:
\begin{itemize}
	\item \emph{Channel--frequency band pruning}: A new technique which prunes
		contiguous frequency coefficients above a given frequency threshold by
		wrapping pointwise convolutions with Discrete Cosine Transforms (DCT).
	\item \emph{FCMask}: A new parameter which can be used to learn the level of
		channel--frequency band pruning per-layer and per-channel during
		training time, short for Frequency Contiguous Mask.
\end{itemize}

\section{Related Work}

Our work builds upon and combines three lines of work on making CNN inference
efficient: one that develops CNN architectures that can be trained from scratch
for small model sizes and fast inference; one that lowers model sizes of an
already trained network by pruning weights; and one that uses frequency-domain
methods to compute CNNs more efficiently, mostly by leveraging the convolution theorem. 
In this section, we describe these three lines of work and explain how they relate to our contributions.

\emph{Efficient CNN architectures.} This paper builds on a line of work on
designing CNN architectures that support efficient inference. One of the
earliest of these architectures was SqueezeNet~\cite{squeezenet}, which achieved
accuracy comparable to a state-of-the-art CNN (AlexNet~\cite{alexnet}) with
substantially fewer weights. MobileNet achieved an even greater reduction in
computation by leveraging depthwise separable convolutions (which were first
introduced in another parameter-efficient architecture,
Xception~\cite{xception}). According to the authors, ``Our model structure puts
nearly all of the computation into dense \onebyone{} convolutions\ldots
MobileNet spends 95\% of it’s computation time in \onebyone{} convolutions which
also has 75\% of the parameters''~\cite{mobilenetv1}. Since then, several other
architectures have been built with similar structure, including
\mobilenet~\cite{mobilenetv2}, ShuffleNet~\cite{shufflenet},
ShuffleNetV2~\cite{shufflenetv2}, and ResNeXt~\cite{resnext}.
EfficientNet~\cite{efficientnet}, EfficientDet~\cite{efficientdet}, and
EfficientNet2~\cite{efficientnetv2} expanded on these architectures by proposing
unified scaling methodologies for increasing network scale and demonstrating
effectiveness when used for object detection and semantic segmentation.

\emph{Weight pruning.} A second approach to efficient inference is
\emph{pruning} in which some of the signals in a neural network are
removed~\cite{prunesurvey}. This can be done at the channel
granularity~\cite{luo2017thinet,he2017channel,zhuang2018discrimination,automl_prune,wang2020pruning} or the weight
granularity~\cite{deepcompression}. While channel pruning results in dense
computation, weight pruning results in sparse computation. Pruning at either
level of granularity results in a smaller model size, but the sparse nature of
weight pruning methods typically results in a significant \emph{increase} in
computation time unless extremely high levels of pruning are
possible~\cite{yu2017scalpel}. Because the goal of this work is to decrease
computation time we have carefully designed our technique to result in dense
computation through the pruning of contiguous (as opposed to arbitrary)
frequency coefficients. Another key difference between traditional pruning
techniques and our technique is that we prune the frequency coefficients of
\emph{activations}, not weights. While this has no impact on model size, we have
demonstrated that it can significantly reduce the amount of necessary
computation.


\emph{Frequency Domain CNNs.} One well-known method to perform convolutions in
the frequency domain is to leverage the Fourier transform's convolution
theorem~\cite{dft_conv}. This method applies the DFT to a padded convolutional
kernel and the input image, element-wise multiplies each frequency component,
and then applies the inverse DFT to the result. While this method can
significantly reduce the total number of operations required, its benefit relies
on large kernel sizes to amortize the cost of the DFT. A closely related
approach is to apply weight pruning in the frequency
domain~\cite{freq_domain_pruning}. This technique suffers from the same sparsity
problem seen in typical weight pruning and is also inapplicable to the pointwise
convolutions which dominate depthwise separable CNNs. Another related technique
in the literature is OctConv~\cite{octconv}. Similar to this work, OctConv
observes that channels within a layer experience heterogeneous frequency
sensitivity. Rather than applying a frequency transformation, however, OctConv
subsamples the subset of channels which are less sensitive to high frequencies.
Our DCT-based approach offers a finer level of granularity in the frequency
domain than OctConv's sub-sampling approach.


\section{Methodology}
\label{sec:meth}

This section describes our technique for computing pointwise convolutions in a
DCT-based frequency space, how pruning can be applied in the frequency domain,
and then how this pruning can be learned with standard back-propagation.

\subsection{Pointwise Frequency Domain Convolution}

Our approach to frequency-domain computation is to compress activation data
using a similar approach as JPEG image compression~\cite{jpeg}: we divide each
channel's activations into small, fixed-size, square tiles called \emph{macroblocks} and apply
the discrete cosine transform (DCT) to macroblock. The key observation is that running
a 1$\times$1 convolution on this transformed data is equivalent to running it on
the original activation:
\[ \text{Conv1$\times$1}(X) = \text{IDCT}( \text{Conv1$\times$1}(
	\text{DCT}( X ))) \]
where $X$ is an activation tensor, DCT is the macroblocked discrete
cosine transform, and IDCT is the inverse transformation. 
Therefore, we can insert DCT and IDCT ``layers'' before and after every
pointwise convoluton and obtain an equivalent network.

Just like JPEG, our technique is sensitive to macroblock size. Where JPEG uses
macroblocks to localize frequency information to specific locations within the
image, in our setting macroblock size represents a trade-off between the
precision of frequency-domain pruning and the computational overhead from
frequency transformation. Larger macroblocks result in more coefficients per
macroblock which offers more precision in the frequency space and thus leads to
better pruning potential. On the other hand, the DCT forms the primary overhead
cost for our technique and the computational cost scales quadratically with the
size of the macroblock. In our experiments, we find that 3$\times$3 macroblocks
work well across both networks we evaluate.

We apply the DCT to each channel separately. To implement the DCT, each
macroblock is multiplied by frequency filters (of size $k^2$ where $k$ is the
kernel width) for each coefficient (of which there are $k^2$). With this
implementation of the DCT, the number of required multiply-accumulate operations is:
\[
	\text{DCT\_MACs} = c_{in} \times \frac{h}{k} \times \frac{w}{k} \times k^4 =
	c_{in} \times h \times w \times k^2
\]
Processing the same input with a pointwise convolution with a stride of 1 would
require the following operations
\[
	\text{Pointwise\_MACs} = c_{in} \times h \times w \times c_{out}
\]
To apply the inverse DCT, each coefficient (of which there are $k^2$) is
multiplied by a kernel (of width $k^2$) to transform back to the spatial domain.
\[
	\text{IDCT\_MACs} = c_{out} \times \frac{h}{k} \times \frac{w}{k} \times k^4
	= c_{out} \times h \times w \times k^2
\]
Because our technique transforms to and from the DCT domain for each pointwise
convolution that is pruned, our overhead consists of the operations required to
do this transformation. The ratio of overhead operations to baseline
pointwise operations is the following for each layer:
\[
	\frac{\text{DCT\_MACs + IDCT\_MACs}}{\text{Pointwise\_MACs}} = 
	\frac{(c_{in} + c_{out})\times k^2}{c_{in} \times c_{out}}
\]
From this overhead equation we can see that the larger $c_{in}$ and $c_{out}$ are
the smaller the transformation overhead is with respect to number of baseline
operations. Typically, the number of channels increases with each subsequent
layer within a CNN and thus the overhead for our technique is smaller
for later layers than earlier layers.

\subsection{Frequency Band Pruning}
\label{sec:pruning}

\begin{figure}[ht] 
  \begin{subfigure}[b]{0.5\linewidth}
    \centering
    \includegraphics[width=\linewidth]{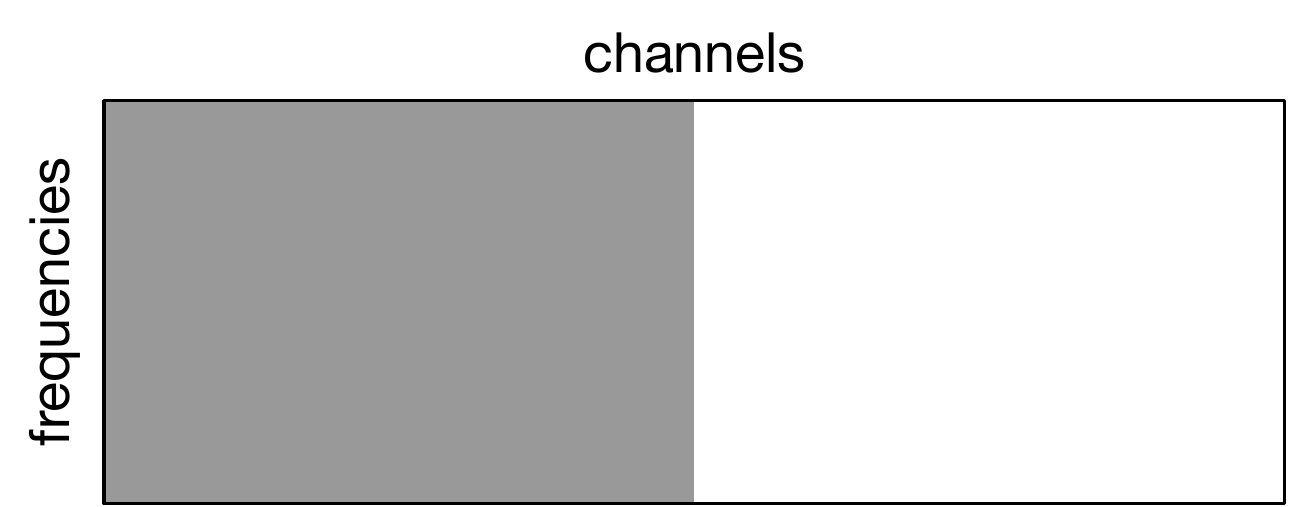}
        \caption{Channel pruning.}
        \label{fig:pruning-chan}
    \vspace{1ex}
  \end{subfigure}
  \begin{subfigure}[b]{0.5\linewidth}
    \centering
    \includegraphics[width=\linewidth]{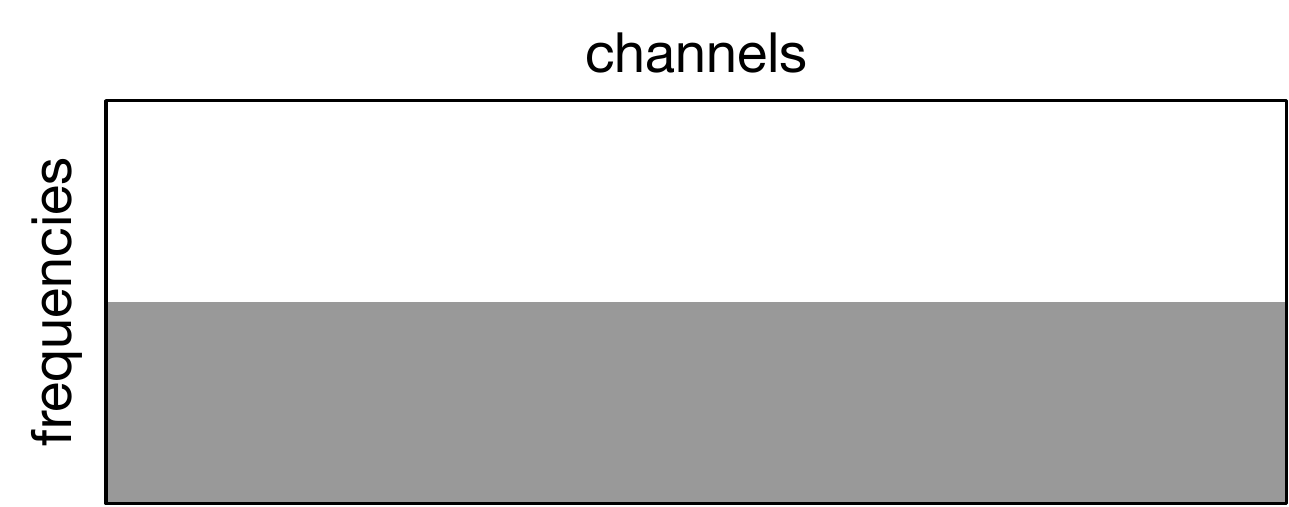}
        \caption{Coefficient pruning.}
        \label{fig:pruning-coeff}
    \vspace{1ex}
  \end{subfigure} 
  \begin{subfigure}[b]{0.5\linewidth}
    \centering
    \includegraphics[width=\linewidth]{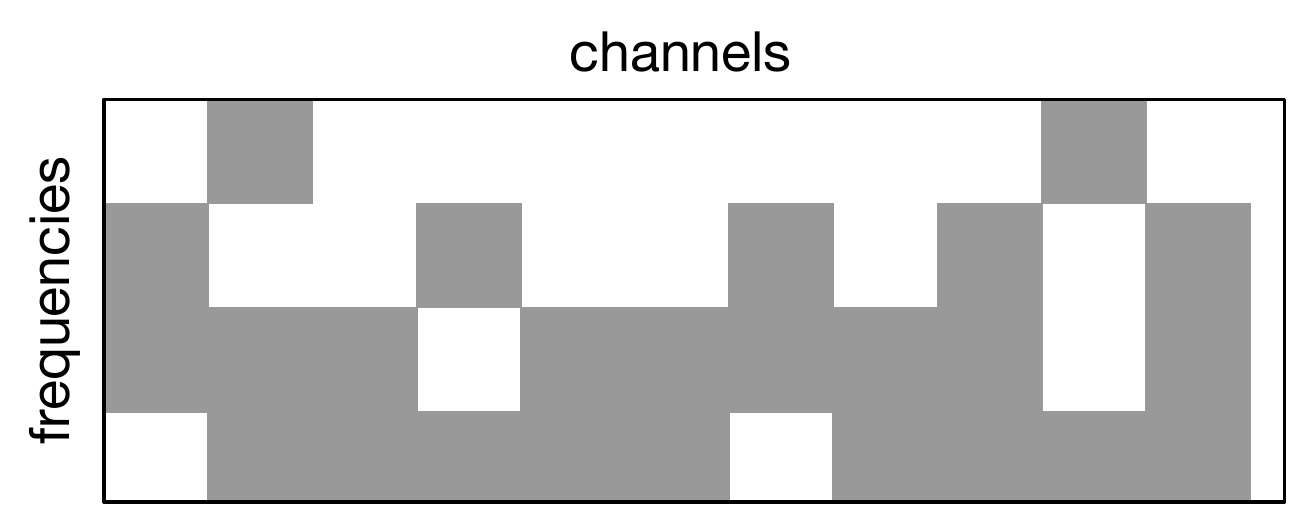}
        \caption{Channel--coef pruning.}
        \label{fig:pruning-chancoeff}
  \end{subfigure}
  \begin{subfigure}[b]{0.5\linewidth}
    \centering
    \includegraphics[width=\linewidth]{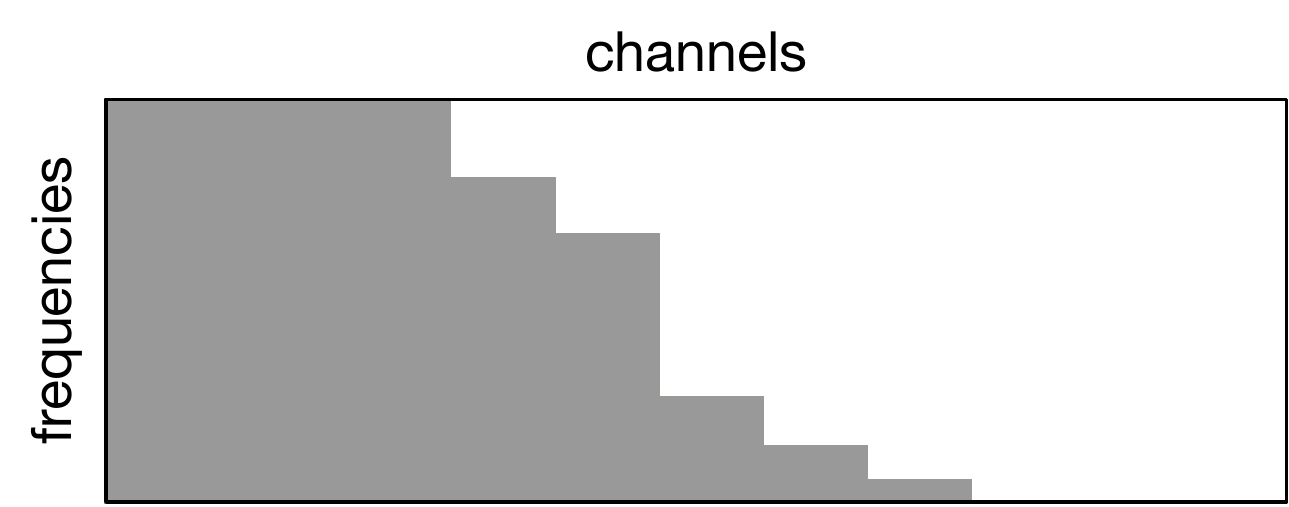}
        \caption{Channel--freq band pruning.}
        \label{fig:pruning-contigband}
  \end{subfigure} 
    \caption{All pruning strategies for frequency-domain \onebyone{}
	convolutions considered in this paper. The frequency axis is according to
	the output of the 2D DCT, vectorized according to the zigzag pattern
	commonly associated with JPEG~\cite{jpeg}. Each figure depicts a possible pruning mask
	for each technique with 50\% pruning. Gray areas are preserved (nonzero) and
	white areas are pruned (zero).}
	\label{fig:pruning}

\end{figure}

Unlike techniques that use the convolution theorem, transforming activations
with the DCT does not save any computation by itself---its purpose is to enable
pruning in the frequency domain. 
We evaluate different pruning techniques shown in Figure~\ref{fig:pruning} building up to our proposed channel--frequency band pruning.

\emph{Channel pruning:} The method we visualize in Figure~\ref{fig:pruning-chan}
removes entire channels and then sorts them such that all non-zero channels are
contiguous. This technique is equivalent to traditional channel pruning methods. This technique prunes on a coarse level since it does not take frequency coefficients into account. However, this allows for an easy channel reordering to allow dense computation.

\emph{Coefficient pruning:} This approach prunes entire coefficients,
independent of the layer's output channels. Figure~\ref{fig:pruning-coeff}
depicts a coefficient pruning mask: all channels preserve the same set of
``important'' frequencies. Coefficient pruning can reduce both the cost of the
pointwise convolution and the overhead associated with the DCT and inverse-DCT
computations as it is not necessary to produce frequency coefficients that have
been pruned. Our technique \emph{truncates} the DCT to produce only the unpruned
coefficients. While this approach results in dense computation, it also
expresses the incorrect assumption that all output channels are \emph{equally
sensitive} to each frequency in the activation data. We find that channels are
\emph{not} uniformly sensitive to frequencies in practice, which motivates a
more nuanced approach.

\emph{Channel--coefficient pruning:} Channel--coefficient pruning removes the
assumption that all channels within a layer are uniformly sensitive to each
frequency. Instead, it acknowledges that some channels need different frequency
information than others. 
Figure~\ref{fig:pruning-chancoeff} depicts the result: each channel receives a
subset of the frequency coefficients.

We find that this per-channel sensitivity significantly increases the amount of
pruning that is possible for a given accuracy budget. However, the additional
flexibility comes at a cost: the ``random'' effect of the pruning means that the
computation must be sparse. It is not generally possible to reorder channels and
coefficients to pack them into a dense tensor for fast computation.

\emph{Channel--frequency band pruning:} Our final approach extends
frequency--coefficient pruning to remove contiguous ranges of frequencies
(bands), thereby recovering dense computation. The key idea relies on the
insight from prior work that CNNs are generally more sensitive to low
frequencies than to high frequencies~\cite{kilian-compress-cnns, deepjpeg,
freq_domain_pruning}. Put differently, coefficient importance is approximately
monotonic with frequency---so if a given coefficient is pruned, all the
higher-frequency coefficients are also likely to be pruned.
\emph{Channel--frequency band pruning} restricts pruning to a contiguous range
of the highest frequencies according to the unique needs of each channel.
Figure~\ref{fig:pruning-contigband} shows a possible mask: each channel receives
a contiguous range of frequencies starting at the lowest frequency.

With this strategy, each channel's computation in the \onebyone{} filter is
dense. We find that this technique prunes nearly the same set of channels as
unrestricted channel--coefficient pruning: significant frequencies naturally
tend to be nearly contiguous. So this technique loses only a few pruned
coefficients while making computation dense and therefore far more efficient.

\subsection{Learned Pruning}

Channel, coefficient, channel--coefficient, and channel--frequency band pruning
each represent different methods for representing coefficient/channel masks, but
how exactly to set the levels for each mask is non-trivial.

A profiling-based selection on a pretrained network was intractable for an exhaustive search. 
To overcome the shortcomings of the profiling-based approach we developed a
new method to learn the best mask for each coefficient using backpropagation.
We created a new learnable parameter called the \emph{frequency
contiguous mask} (or FCMask) which represents the level of pruning for each
channel while also maintaining frequency contiguity. To apply pruning, each
coefficient in each channel is multiplied by a coefficient mask value derived from
the FCMask value for that channel.

\begin{figure} \centering
	\includegraphics[width=0.6\linewidth]{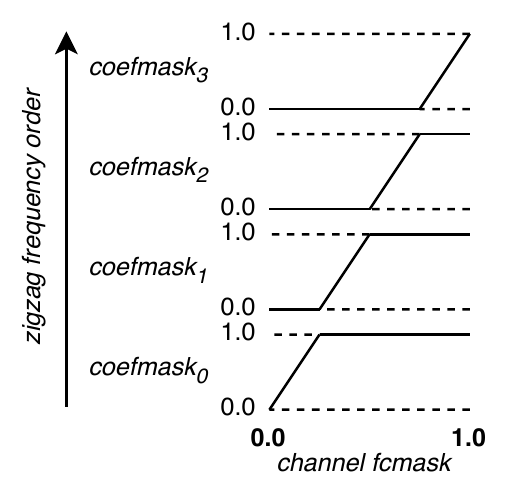}%
	\caption{Here we graph how each individual coefficient's mask is a
	function of FCMask. This example is for four coefficients, which would
	correspond to a 2$\times$2 macroblock}
	\label{fig:learned_freqcontigmask}
\end{figure}
\begin{equation*}\label{eq:coefmask_FCMask}
  \begin{aligned}
	  \text{num\_coefs} &= \text{macrobock\_width}^{2} \\
	  \text{coefmask}_n &= \left(\text{FCMask} - \frac{n}{\text{num\_coefs}} \right) \times \text{num\_coefs}
  \end{aligned}
\end{equation*}
As Figure~\ref{fig:learned_freqcontigmask} shows, each coefficient's mask is a function
of the corresponding channel's FCMask. When
FCMask is set to 0.0 all coefficient masks in the channel are equal to 0.0
(removing all data from the channel entirely) and when FCMask is set to 1.0 all
frequency masks are equal to 1.0 (no pruning is applied). FCMasks are
initialized to 1.0 and are influenced to decrease by our modified loss function.
Consider the case where a channel's FCMask gradient decreases the value of
FCMask from 1.0 to 0.9. Coefficient masks 0-2 will remain at 1.0, but frequency
mask 3 will slightly decrease in value. Because during inference each
coefficient value is multiplied by its corresponding coefficient mask, the
decrease in the coefficient mask will decrease the impact that this individual
coefficient has on the network as a whole. It is this ability to make small
adjustments to the impact of each coefficient that enables this learning
technique to succeed, since rounding all masks to 0 or 1 during training would
prevent gradients from propagating through the network. After training is
completed, all coefficient masks are rounded and fixed to 0 or 1 so that we can
save on computation.

The modified loss function which influences FCMask values to be decreased can be seen below:
$$
\text{loss} = \text{cross\_entropy} + \lambda \times \sum_{l=0}^{\text{num\_layers}} \text{avg}(|\text{FCMask}_{l}|)
$$
We preserve a cross entropy loss component to maintain the accuracy of the
classification task that our model is trained for. We also introduce a new
metaparameter, $\lambda$, to control the degree of impact that the new loss
component will have on the network as a whole. The primary regularization value
in this loss function is a sum taken over all layers of the average absolute
value for FCMasks over all channels in each layer. Smaller FCMask values (increased pruning)
decrease the loss function and are thus rewarded. On the other hand, masking
coefficients may increase the cross entropy if these coefficients are important
to the network. We have intentionally built in this tension into the loss
function to achieve a balance between efficiency and accuracy.


\begin{figure}
    \centering
        \includegraphics[width=\linewidth]{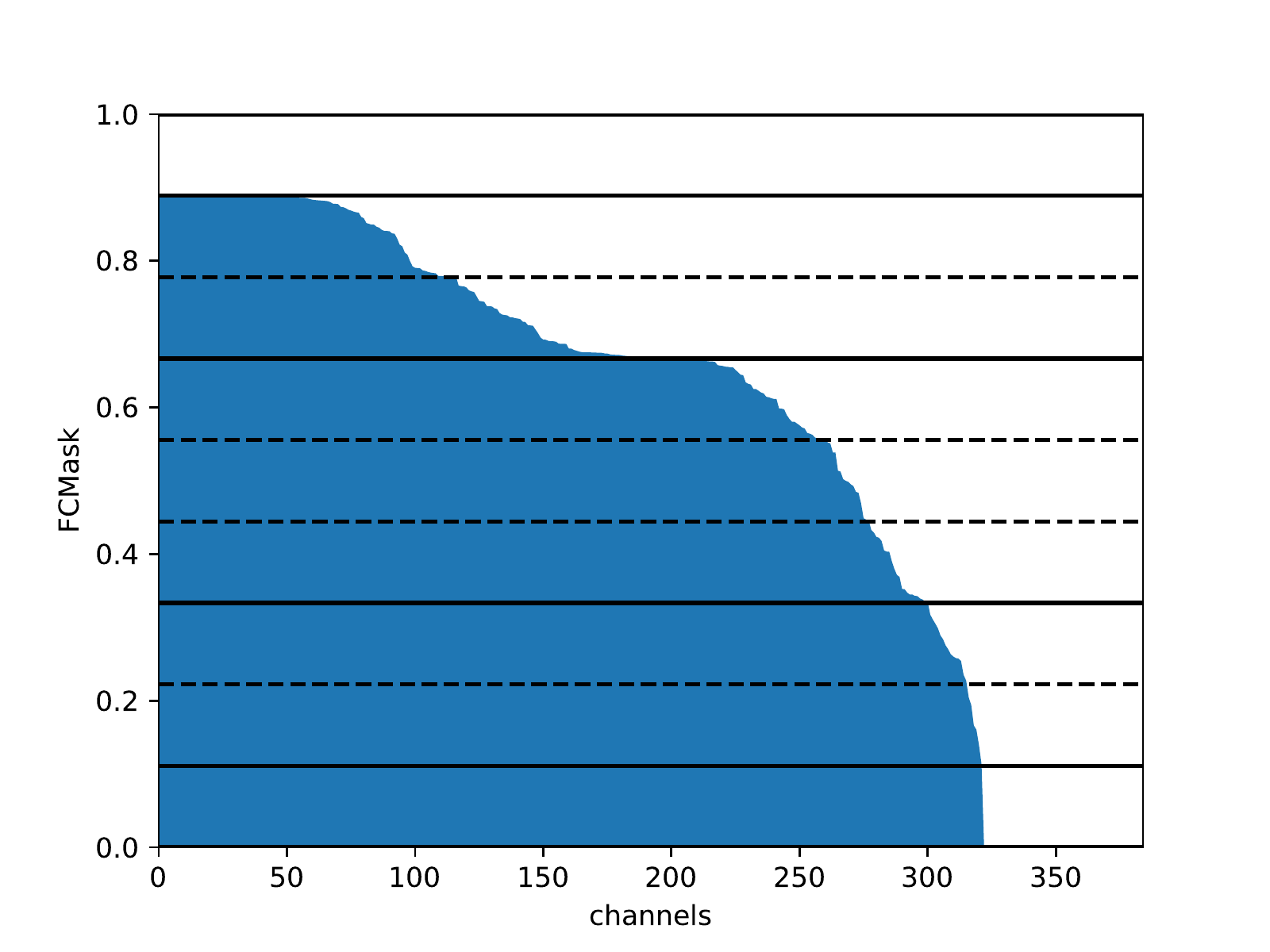}%
    \caption{FCMask values for all channels (sorted from high FCMask to low) in
        the first pointwise layer of Block 10 of \mobilenet using a 3$\times$3
        macroblock size. The horizontal lines demarcate the regions of FCMask
        values which pertain to each of the nine coefficients. The solid
        horizontal lines indicate a transition between diagonal rows in the
        DCT, and the dashed horizontal lines indicate a transition within a
        diagonal row.}
    \label{fig:learned_freqcontigmask_example}
\end{figure}

Figure~\ref{fig:learned_freqcontigmask_example} shows an example of a learned
frequency contiguous mask for MobileNet v2 when trained on ImageNet. The
heterogeneous sensitivity of each channel to frequency can be observed as there
is significant variation in the value of FCMask for each channel.
Figure~\ref{fig:learned_freqcontigmask_example} shows that this layer has has
pruned away the highest frequency coefficient for all channels, and has
also pruned all coefficients for a subset of channels.

\subsection{Dense Frequency-Domain Computation}

\begin{figure} \centering
	\includegraphics[width=0.9\linewidth]{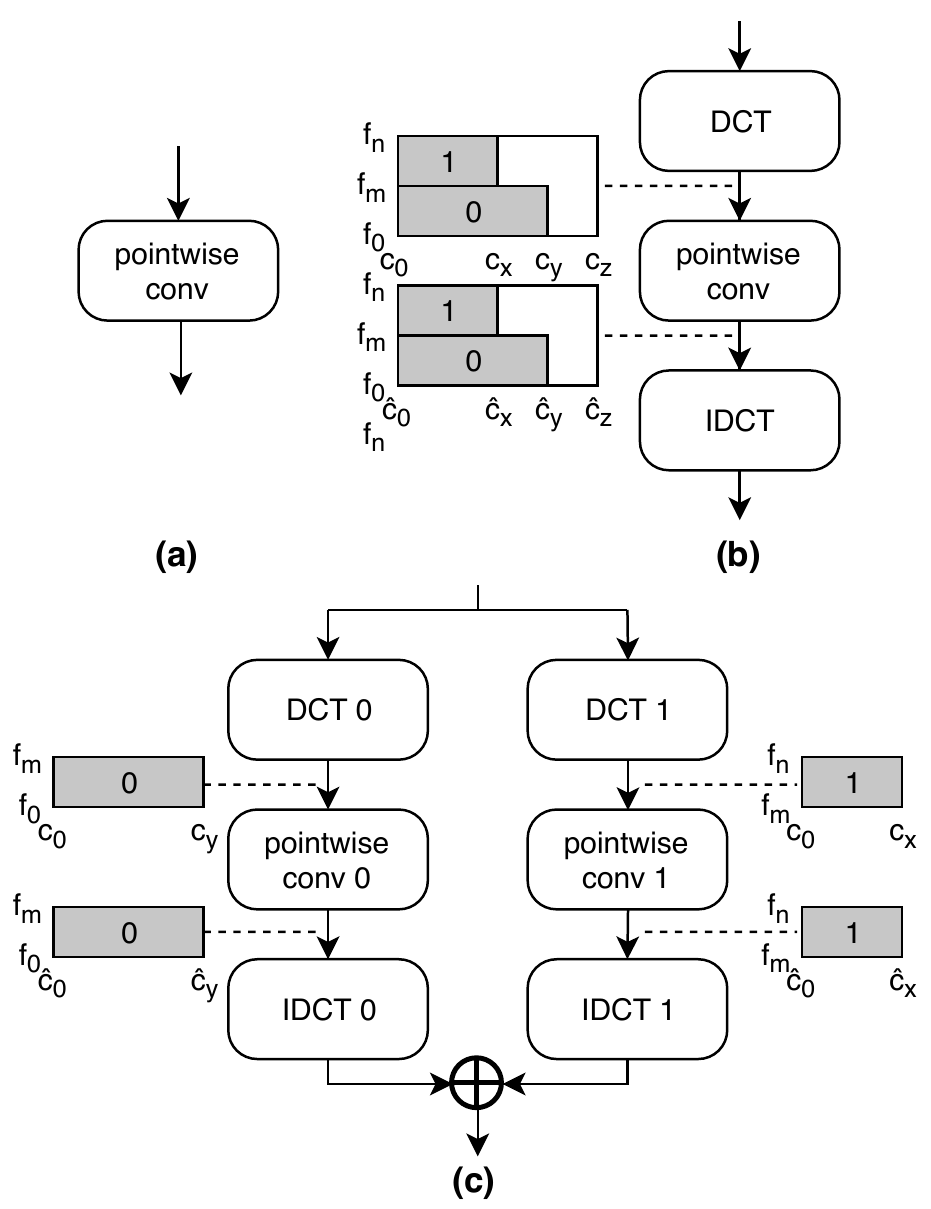}%
	\caption{Three equivalent pointwise convolutions:
	typical, frequency domain, separate frequency bands}
	\label{fig:dense_compute}
\end{figure}

The primary motivation for our pruning technique over traditional pruning is
that our method enables dense computation while still requiring fewer operations
than an unpruned convolution. Figure \ref{fig:dense_compute}(a) shows a
standard pointwise layer, and (b) shows a mathematically equivalent
operation which applies the DCT before and IDCT after the pointwise operation.
The DCT and IDCT masks in Figure \ref{fig:dense_compute}(b) shows that not all coefficients are
needed for all channels, but (b) would require expensive sparse computation to
avoid computing the pruned coefficients.

Figure~\ref{fig:dense_compute}(c) shows our proposal: group coefficients into
frequency bands and then compute the pointwise convolution on each frequency
band separately. The baseline pointwise convolution would
need to multiply all weights by all coefficients in each input channel to
produce all coefficients for all output channels. In contrast,
dividing the computation into frequency bands offers the ability to only process
the channels that pertain to the frequency band of interest. In the example
shown in Figure \ref{fig:dense_compute}(c), frequency band 0 only has non-zero
coefficients from channels $c_0-c_y$ at the input, and only produces non-zero
output for channels $\hat{c}_0-\hat{c}_y$. Similarly, frequency band 1 only has non-zero
coefficients from channels $c_0-c_x$ at the input, and only produces non-zero
output for channels $\hat{c}_0-\hat{c}_x$. By reducing the total volume of input
and output data for each pointwise convolution we have saved a significant
amount of computation while remaining entirely dense.

\section{Evaluation}
\label{sec:exp}

We evaluate the effectiveness of our frequency band activation pruning on two
state-of-the-art networks, \mobilenet~\cite{mobilenetv2} and
\resnext~\cite{resnext}. We begin by performing experimentation on the
CIFAR-10~\cite{cifar10} classification dataset, and then apply the best
performing technique to \mobilenet when trained on ImageNet~\cite{imagenet}. We
evaluate computation savings in two ways. First, we measure the number of
multiply--accumulate (MAC) operations among all convolutional layers. Second,
we measure the wall-clock time to perform all network computation when
performing \mobilenet on ImageNet.

\subsection{CIFAR-10}

This section contains our preliminary experiments to determine the
effectiveness of each of the different frequency domain pruning methodologies.
Figure~\ref{fig:resnext_cifar10} and Figure~\ref{fig:mobilenetv2_cifar10} show
the accuracy vs pruning level tradeoff, although these sweeps do not include
retraining after pruning. Table~\ref{table:highlvl_results} contains the
retraining results after pruning ResNeXt and MobileNetV2 with the most
effective technique: learned channel-frequency band pruning.

In the CIFAR-10 baseline, \resnext uses $7.7\times10^{8}$ MACs
and \mobilenet uses $8.9\times10^{7}$. We add DCT and inverse-DCT transformation
layers to each using 4$\times$4 macroblocks, which increases the MAC counts to
$8.4\times10^{8}$ for \resnext and $1.1\times10^{8}$ for \mobilenet. The top-1
CIFAR-10 accuracy for the baseline, unpruned configuration is 94.2\% for
\mobilenet and 95.75\% for \resnext.

\begin{table}[]
	\centering
	\begin{tabular}{lllll}
	\toprule
					  & \resnext & \mobilenet \\ \midrule
	Acc Degradation   & 0.13\%      & $-0.07$\%   \\
	MAC Reduction     & 2.4$\times$ & 2$\times$ \\ \bottomrule
	\end{tabular}
	\caption{High-level CIFAR-10 results for top-1 accuracy degradation
	(after refinement training) and computational
		savings}
	\label{table:highlvl_results}
\end{table}

Table~\ref{table:highlvl_results} summarizes the results from our evaluation for
our preferred pruning technique, per-channel frequency band pruning. This
technique achieves 2--2.4$\times$ computational savings, as measured by the
number of multiply--accumulate (MAC) operations, with reductions in top-1
accuracy below one fifth of a percent. These numbers result from first training
each network from scratch, 
learning frequency coefficients to prune while keeping the original network weight constant,
and then refining each network with another round of training.

We evaluate each of the pruning techniques in Section~\ref{sec:pruning} by
sweeping a threshold that masks out a given percentage of operations for each
transformation layer. These new pruning techniques are coefficient pruning,
channel--coefficient pruning, and per-channel frequency band pruning. We also show the learned technique for per-channel frequency band pruning. We compare
against two baselines: channel pruning and reducing the input resolution.


\subsubsection{\resnext}
\label{sec:eval:resnext}

\begin{figure}
	\centering
    \includegraphics[width=\linewidth]{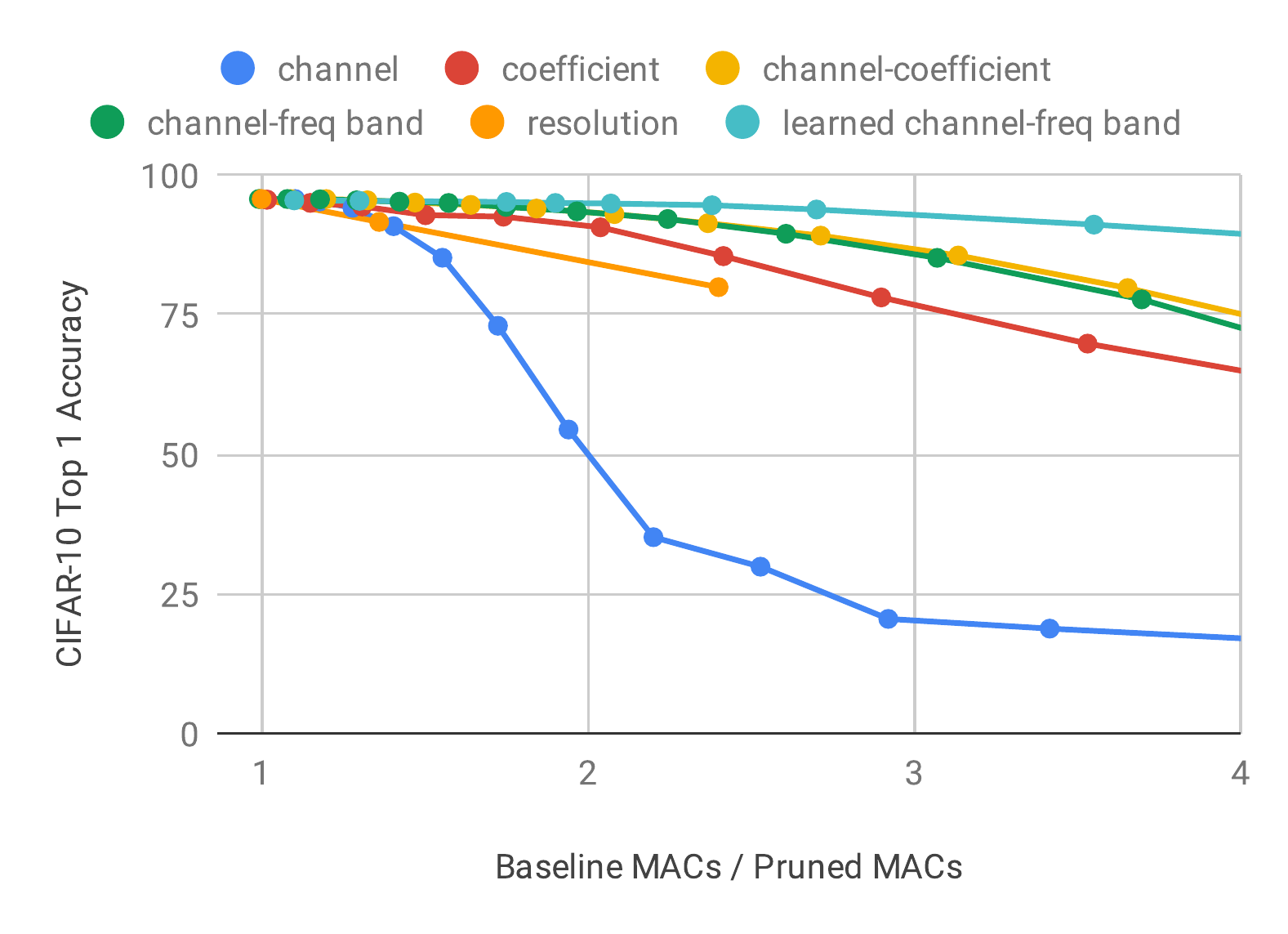}%
    \caption{Comparison of pruning methods for \resnext on CIFAR-10 (before refinement training).}
    \label{fig:resnext_cifar10}
\end{figure}

In \resnext, the \onebyone{} convolutions account for 85\% of all MACs in the
network. The large number of channels per layer (64--1024) means that the DCT
overhead is proportionally smaller when compared with \mobilenet.
Figure~\ref{fig:resnext_cifar10} shows the accuracy for each pruning method at
different levels of pruning by relating the resulting MAC reduction with the
network's top-1 CIFAR-10 accuracy. Channel pruning performs very poorly compared
with all other techniques, demonstrating that the frequency dimension is
important. Also, all frequency-based techniques outperform a simple reduction in
input resolution (input\_res in the figure), suggesting that it is valuable to
remove some but not all of the high-frequency data in the image.

Channel--coefficient pruning, being the most granular pruning technique performs the best compared to other profiling based pruning methods. As Section~\ref{sec:meth} discusses, this fine pruning granularity comes at the cost of sparse computation.

%
Fortunately, the final technique, per-channel frequency band pruning, performs
nearly as well while yielding dense computation. Below 2$\times$ MAC savings,
the two techniques are nearly identical. The similarity between unrestricted
channel--coefficient pruning and the frequency band equivalent validates our
claim that sensitivity is monotonic with frequency. With unrestricted
channel--coefficient pruning at a pruning percentile of 30\% (1.8$\times$ MAC
savings), we measure that 98.8\% of channel masks in all layers of \resnext were
\emph{already} contiguous.

The learning based technique applied to the channel-frequency band pruning
technique performs the best. This is achieved by adding our FCMask
regularization to the training loss, allowing heterogeneous pruning across
layers without signficantly impacting the task accuracy after retraining. The learned technique far exceeds the performance of even
channel--coefficient pruning when profiling.


\subsubsection{\mobilenet}
\label{sec:eval:mobilenetv2}

\begin{figure}
	\centering
    \includegraphics[width=\linewidth]{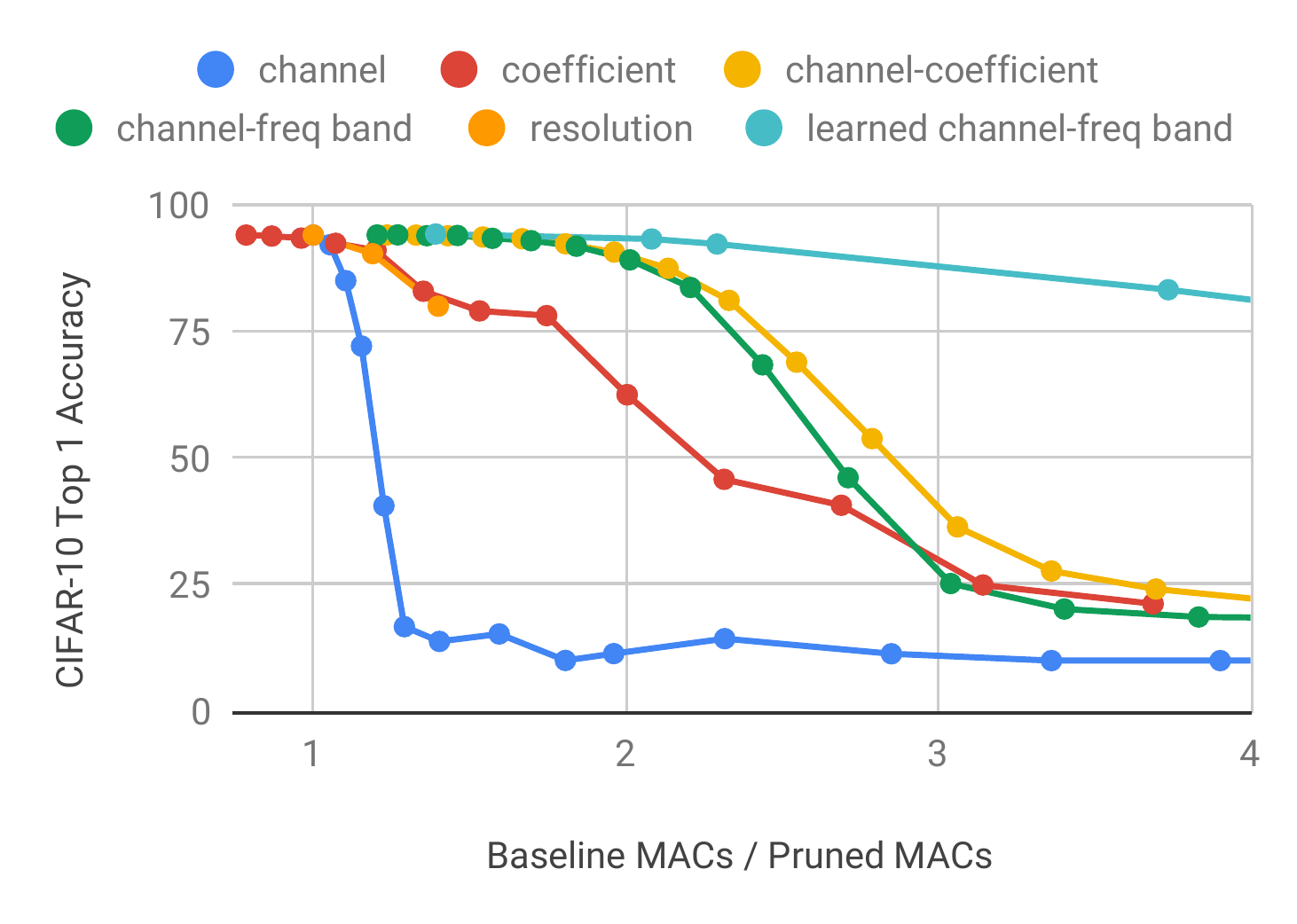}%
    \caption{Comparison of pruning methods for \mobilenet on CIFAR-10 (before
		retraining).}
    \label{fig:mobilenetv2_cifar10}
\end{figure}

We apply our technique to \mobilenet to examine its effects in the context of  a
network that is already highly efficient. Figure~\ref{fig:mobilenetv2_cifar10}
shows our techniques' performance alongside baseline techniques. The
highest degree of per-channel frequency band pruning achieves a
1.8$\times$ reduction in total network MACs.
With retraining, the network exceeds the baseline top-1 accuracy by 0.05\%.

As the figure's $x$-axis scale reveals, \mobilenet is less robust to
pruning overall. The network has already been heavily compressed by design,
leaving less opportunity for removing ineffectual computations. However, per
channel coefficient pruning and per channel frequency band pruning maintain
their record as the best available pruning methods, and all frequency based
methods again outperform reducing input resolution.

In Figure~\ref{fig:mobilenetv2_cifar10}, coefficient pruning starts with
0.86$\times$ MAC savings---i.e., it uses \emph{more} MACs than the baseline. In
this configuration, the 4$\times$4 DCT overhead is significant in the context of
a network with fewer channels per layer (16--320). Neither channel pruning nor
input resolution changes require DCT transformations, so they both incur no MAC
overhead.

In contrast, channel--coefficient pruning and its frequency band variant start
with \emph{fewer} MACs than the baseline, yielding about 1.2$\times$ savings in
their initial configurations. These initial savings arise because, after
applying the DCT, many of the frequency coefficients are zero already and
require no thresholding. In fact, the output of the first \onebyone{} layer and
the input to the second \onebyone{} layer of each residual block exhibit
40--65\% sparsity in the frequency domain. This high degree of sparsity arises
from the sparsity of the depthwise 3$\times$3 filters themselves when in the
frequency domain. Unlike the grouped 3$\times$3 filters of \resnext, where any
of the input kernels in a group may be sensitive high frequencies and therefore
cause the output activations to express high frequencies, each depthwise
convolutional layer's output is only a function of one input channel and one
kernel. This increased granularity in the kernel space improves our ability to
prune.

The learning based technique outperforms the other profiling based techniques, specifically at speedups greater than 2$\times$. This is consistent with \resnext results on CIFAR-10 but the effect is even more significant with \mobilenet since it is a much smaller network.

\subsection{ImageNet}
\begin{figure} \centering
	\includegraphics[width=\linewidth]{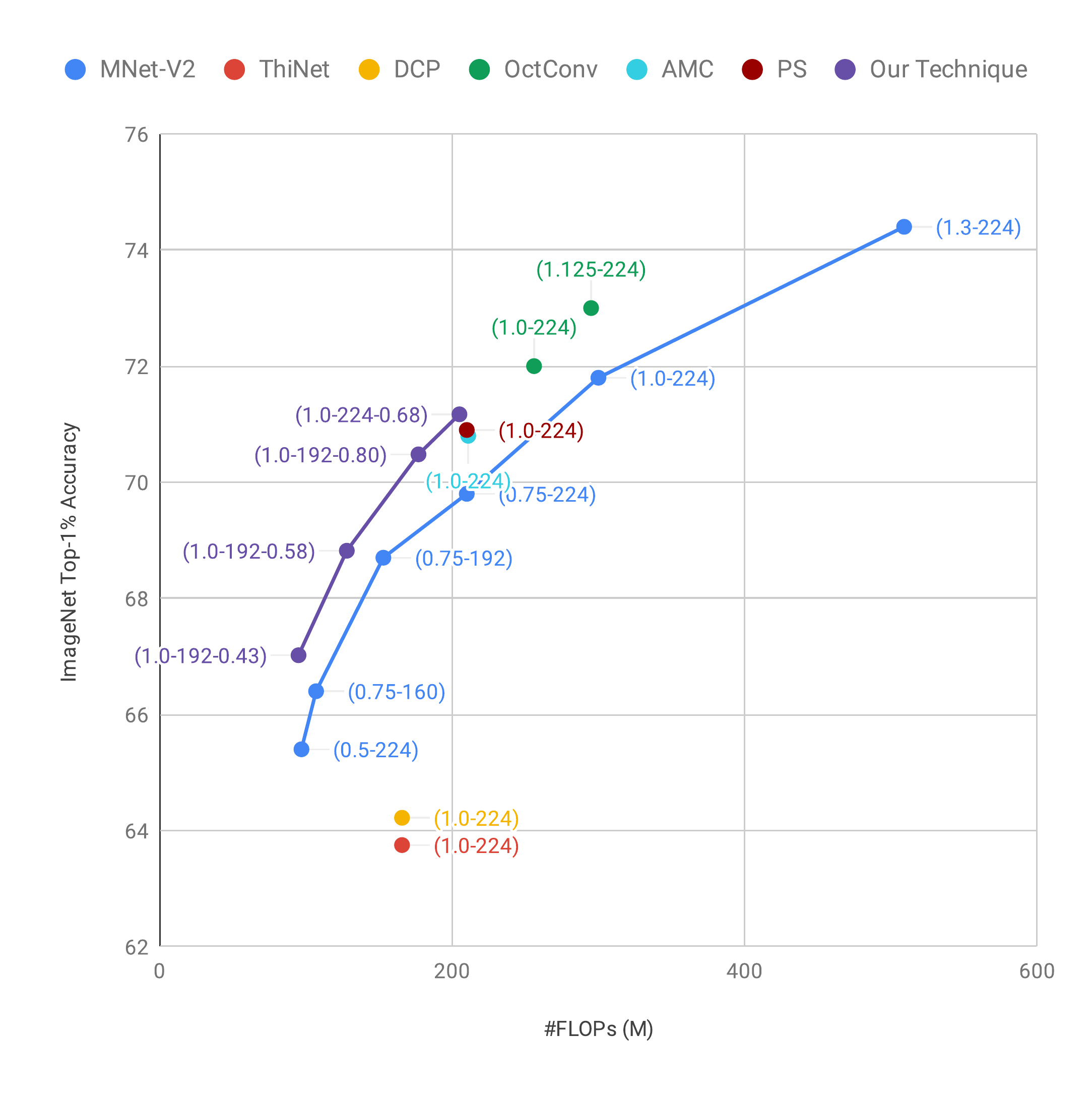}%
	\caption{\mobilenet Pruning Method Effectiveness on the ImageNet Dataset. Each point is labeled with parentheses, where the first number refers to the width multiplier and the
	second refers to the input resolution. For our technique, we also add a pruning ratio to denote multiple compression levels for the same \mobilenet configuration.}
	\label{fig:mobilenetv2-imagenet}
\end{figure}

\begin{figure*}
	\includegraphics[width=\textwidth]{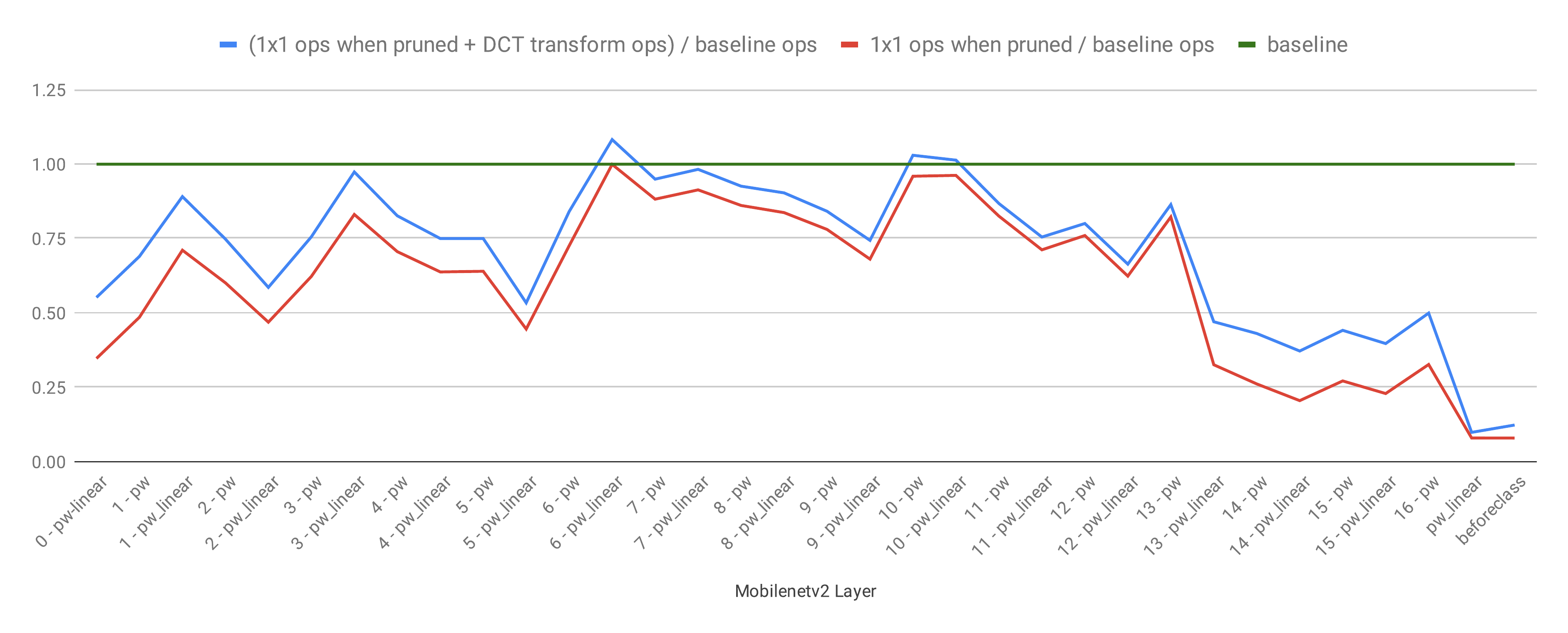}%
	\caption{\mobilenet Layer-Wise Frequency Coefficient Pruning for (1.0-224) configuration with 205M FLOPs (0.68$\times$ FLOP reduction) on ImageNet. The red line is the MAC compression in \onebyone convolutions and the blue line is the overall compression including the transform overheads.}
	\label{fig:layer-wise-pruning}
\end{figure*}

\begin{table}[]
	\centering
	\footnotesize
	\begin{tabular}{lllll}
	\toprule

	\multicolumn{2}{c}{Model} 		& AMC\cite{automl_prune} & PS\cite{wang2020pruning} & Ours \\\midrule
	\mobilenet 		  & MACs  	  	& 211M & 210M & 205M \\
	(Baseline 71.8\%) & Top--1 Acc. & 70.8\% & 70.9\% & \textbf{71.17}\% \\\bottomrule
	\end{tabular}
	\caption{ImageNet top-1 accuracy comparison with state of the art compression models with comparable pruning levels. }
	\label{table:highlvl_imagenet}
\end{table}

Table~\ref{table:highlvl_imagenet} summarizes our results for \mobilenet
evaluation on ImageNet comparing with other state-of-the-art models which
compress the same network resolution to comparable pruning levels. Our method is able to prune the \mobilenet model to similar levels of 205M MACs while achieving the best top-1 accuracy.

Figure \ref{fig:mobilenetv2-imagenet} compares our technique's ability to reduce MAC operations
on varying resolutions of \mobilenet when compared with other methods with
varied pruning levels.
The figure shows the accuracy and operation count for multiple
configurations of \mobilenet which forms a Pareto curve. We compare our technique to recent
channel pruning models: ThiNet~\cite{luo2017thinet}, Discrimination-aware
Channel Pruning (DCP)~\cite{zhuang2018discrimination}, Automatic Model
Compression (AMC)~\cite{automl_prune} and Pruning from Scratch
(PS)~\cite{wang2020pruning}. We also plot OctConv~\cite{octconv} which introduces a convolution operator using a combination of low- and high-resolution feature maps. While OctConv is not strictly a pruning method, it does leverage features' heterogeneous sensitivity
to different frequencies via multi-resolution feature maps. 
The blue pareto curve is the baseline \mobilenet architecture with varying
input resolution and network width scaling.  The violet curve represent points
with different input resolution and varying pruning levels of our learned
technique. After learning the frequency pruning, we train the network weights
from scratch to allow channels to rearrange weights based on the learned
frequency importance. We typically see an improvement of 1--2\% in top-1
accuracy compared to a simple refinement strategy. We used a $3\times3$
macroblock for the $192\times192$ resolution and a combination of $2\times2$
and $7\times7$ macroblocks for $224\times224$ resolution for our ImageNet
experiments.


Our technique compares favorably with ThiNet, DCP, AMC, PS channel compression,
and OctConv as our technique unifies the idea of
pruning and tuning channel sensitivity to frequency. Our
channel--frequency band pruning can both prune entire channels and reduce spatial redundancy through frequency pruning. The comparison to OctConv is not straightforward since our technique does offer more frequency pruning precision than simple subsampling, but this additional precision comes at the computational cost of frequency transformation. Figure~\ref{fig:layer-wise-pruning} shows the impact our transformation overhead for each layer.  The earlier layers have higher resolution input activations and the later layers have much higher number of channels, both of which can be effectively compressed by our learned pruning technique. Since we use a smaller $2\times2$ macroblock for the layers up to block 13, we get less pruning than the later layers with $7\times7$ macroblocks. However the overhead is higher for the layers with $7\times7$ macroblocks, evident by the gap between the two lines. This tension between macroblock sizes and overhead along with the hyperparameter tuning of the loss function allows us to learn fine-grained pruning levels as an independent scaling dimension to input resolution and width multiplier.

\subsection{Wall-Clock Speedup}

The purpose of dense pruning in the frequency domain is to offer speedups in actual execution time, not just reduction in total operation count, over the baseline network.


We implement the (1.0-192-0.80) \mobilenet configuration from Figure~\ref{fig:mobilenetv2-imagenet} in TVM. The 20\% FLOP compression is achieved by only applying pruning to those layers where the pruned \onebyone operations + transform overhead is smaller than the baseline FLOPs. We measured the time taken with and without our technique on an AWS c5.18xlarge instance with Xeon Platinum 8124M CPU and 140G DRAM, running on single thread. When we apply our technique resulting in the 20\% FLOP reduction, the resulting network is 22.1\% faster, while incurring \textless1\% accuracy degradation.




\section{Implementation Details}
\label{sec:impldet}

Our \resnext CIFAR-10 model uses a cardinality of 32 and a bottleneck width of
4, while our \mobilenet CIFAR-10 model uses a width multiplier of 1.0. We built
and trained our networks using PyTorch and four Nvidia GTX Titan X GPUs.
Training hyperparameters are the same as those used in the original papers.




\section{Conclusion}

This paper does two things that at first seem paradoxical: it prunes CNNs
without introducing sparsity, and it applies frequency-based methods to save
computation on kernels with no spatial context. Beyond simply saving
computation in theory we have achieved real-world speedup.

{\small
\bibliographystyle{ieee_fullname}
\bibliography{egbib}
}

\clearpage
\appendix
\section{Appendix}

\subsection{CIFAR-100 experiments}

In the CIFAR100 baseline, \mobilenet uses $4.3\times10^7$ MACs. We add DCT and inverse-DCT transformation layers using $4\times4$ macroblocks, which accounts for 20\% overhead and increases the total MAC count to $5.4\times10^7$. The top--1 accuracy for \mobilenet unpruned baseline configuration on CIFAR100 is 67.92\%. 

\begin{table}[h]
	\centering
	\begin{tabular}{llll}
	\toprule
                      & \mobilenet \\
                      & (Baseline 67.92\%) \\ \midrule
	Top--1 Acc        & $67.04\%$   \\
	MAC Reduction     & 3.77$\times$ \\ \bottomrule
	\end{tabular}
	\caption{High-level CIFAR-100 results for top-1 accuracy (after refinement training) and computational savings}
	\label{table:highlvl_results_c100}
\end{table}

Table~\ref{table:highlvl_results_c100} summarizes the results from our evaluation for per-channel frequency band pruning. This technique achieves 3.77$\times$ computational savings, as measured by the number of multiply--accumulate (MAC) operations, with less than a percent of top--1 accuracy reductions. These numbers result from first training
each network from scratch, learning frequency coefficients to prune while keeping the original network weight constant, and then refining each network with another round of training.

We also experimented different learning schemes for \mobilenet on CIFAR-100 dataset. CIFAR-100 enabled us to do fast experiments while not overtraining on small datasets like CIFAR-10. The two learning methods we tried are:
\begin{itemize}
    \item Learn mask followed by weight refinement (baseline): We freeze weights and learn masks for the trained network. Once we achieve the desired pruning levels, we freeze the pruning and refine the weight to improve accuracy.
    \item Alternate between learning mask and refining weights: This method alternates between learning masks and refining weights in each epoch. We can allow the network to prune more aggressively while maintaining accuracy.
\end{itemize}

\begin{figure}
	\centering
    \includegraphics[width=\linewidth]{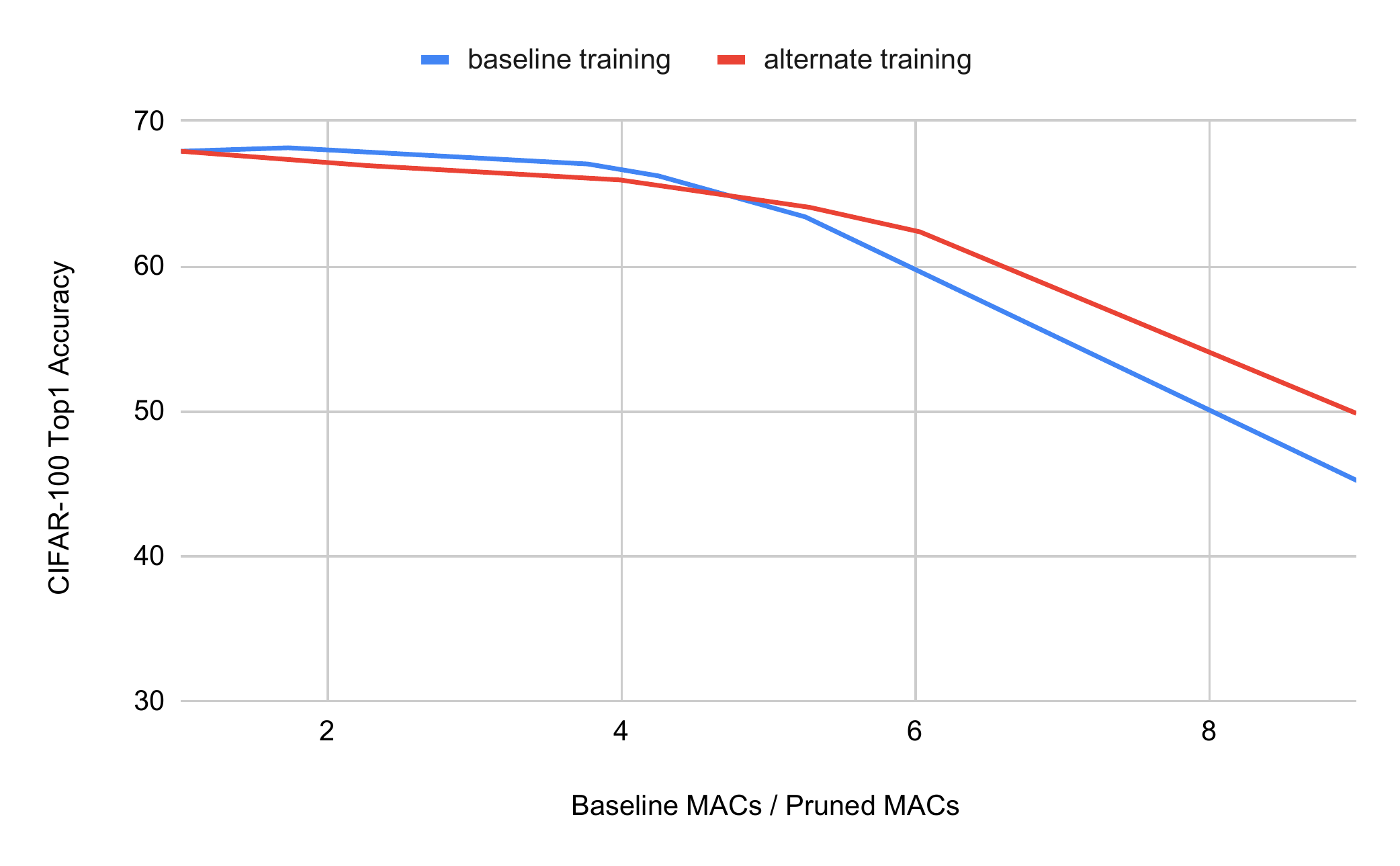}%
    \caption{Comparison of training schemes for \mobilenet on CIFAR-100.}
    \label{fig:cifar100_alt}
\end{figure}

Figure~\ref{fig:cifar100_alt} shows the Top--1 accuracy on CIFAR-100 using the two training methods for different pruning levels resulting in the overall speedup. While running experiments we noticed that it was easier to train for higher pruning levels with alternate training method, since it allowed refinement in regular intervals and did not allow the network accuracy to tank. This is evident from the graph as alternate training performs better at higher speedups. The slight decrease in accuracy for alternate training for lower speedups can be attributed to freezing the batchnorm values while training and refining. This was rectified during Imagenet experiments by allowing batchnorm to finetune toward the end.

\begin{figure}
	\centering
    \includegraphics[width=\linewidth]{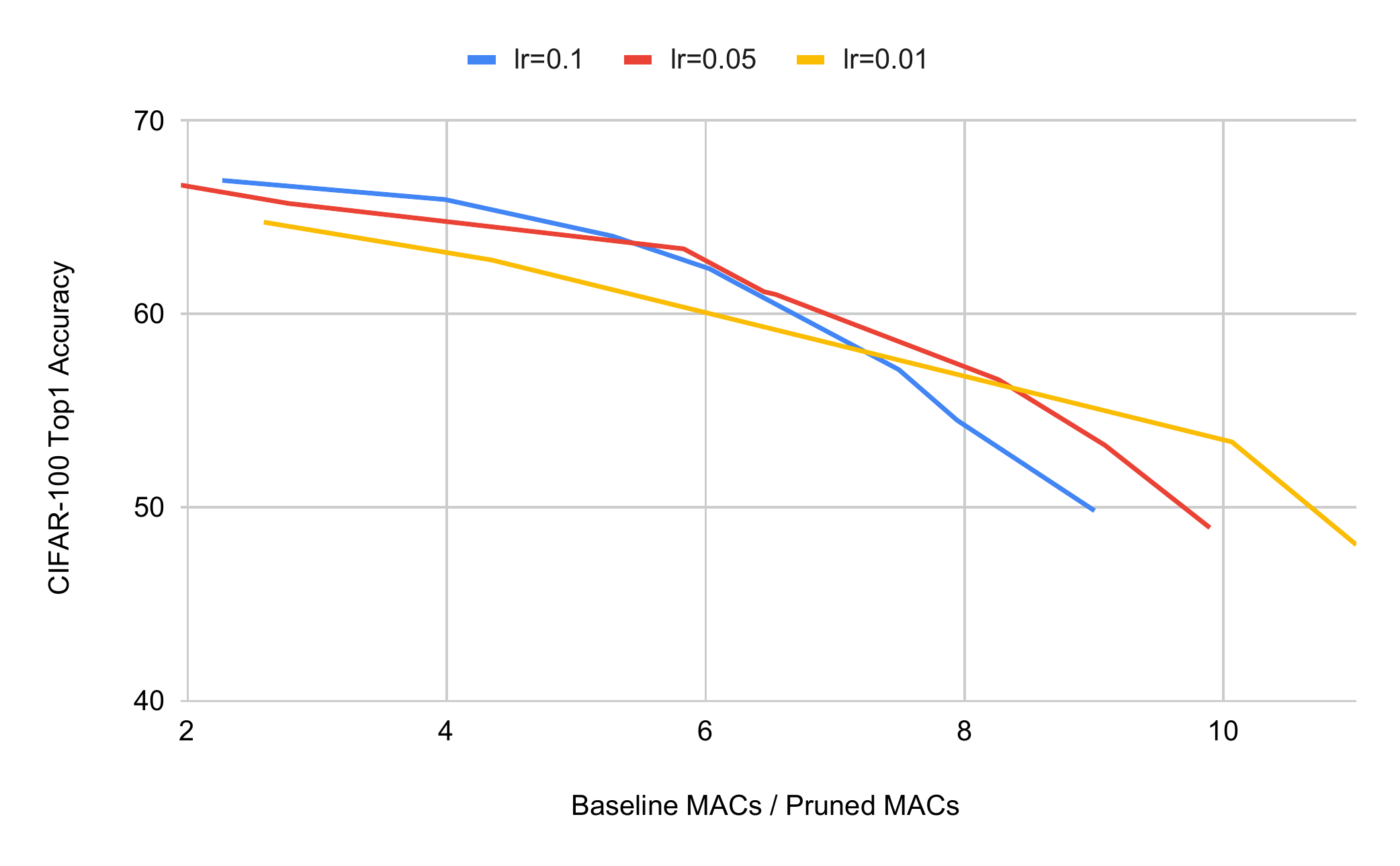}%
    \caption{Hyperparameter tuning in alternate training for \mobilenet on CIFAR-100. We can select different learning rates based on the desired speedup/pruning level.}
    \label{fig:cifar100_lr}
\end{figure}

Figure~\ref{fig:cifar100_lr} shows the different hyperparameter tuning results for \mobilenet on CIFAR-100. We observe that the developer can choose different learning rates based on the speedup desired from the model.

\begin{figure*}[t!]
    \centering
    \begin{subfigure}[t]{0.5\textwidth}
        \centering
        \includegraphics[width=\linewidth]{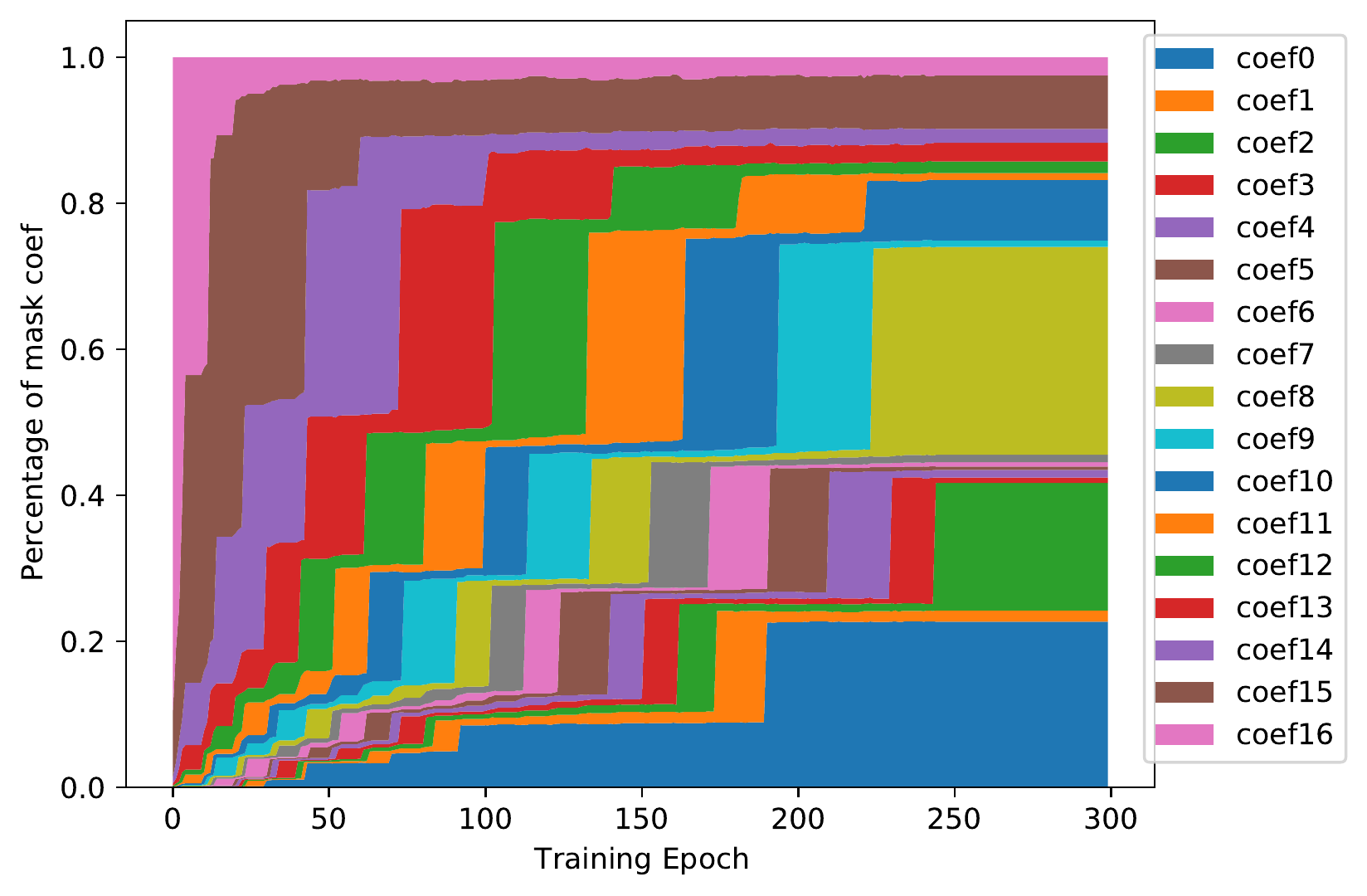}
        \caption{\mobilenet trained on CIFAR-100 with $2.27\times$ MAC reduction}
        \label{fig:a:cifar100_coef_pruning}
    \end{subfigure}%
    ~~~
    \begin{subfigure}[t]{0.5\textwidth}
        \centering
        \includegraphics[width=\linewidth]{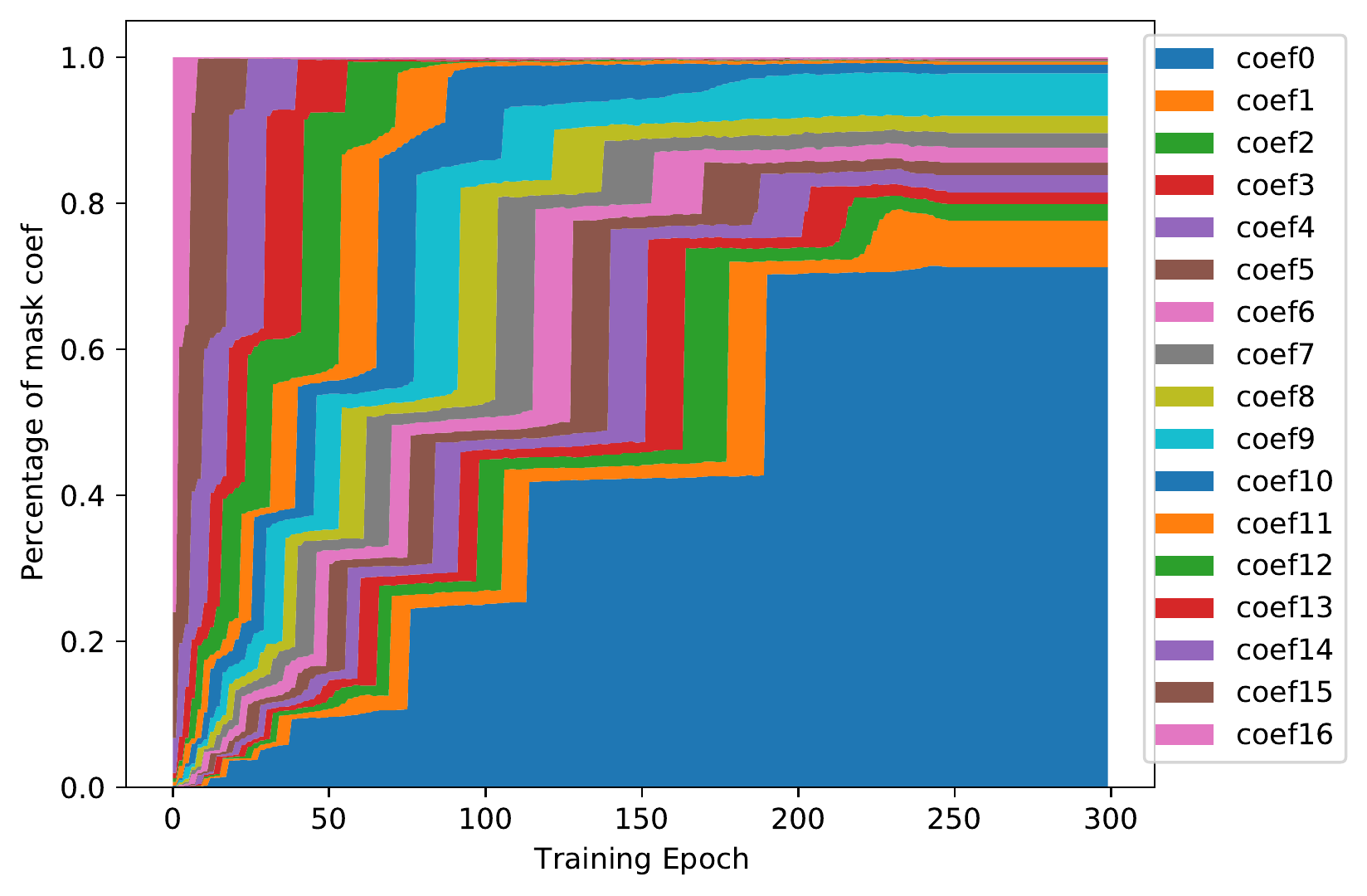}
        \caption{\mobilenet trained on CIFAR-100 with $5.83\times$ MAC reduction.}
        \label{fig:b:cifar100_coef_pruning}
    \end{subfigure}
    \caption{Learning to prune weights in the frequency domain for \mobilenet. The coefficients represent pruning starting from that level till the highest frequency coefficient. Since we use $4\times4$ macroblocks, we have 16 coefficients. "coef0" represents the channels which prune away all frequency components and "coef16" represents those which do not prune anything.}
    \label{fig:cifar100_coef_pruning}
\end{figure*}

Figure~\ref{fig:cifar100_coef_pruning} shows different levels of fine-grained frequency pruning existing in the network and the progression of learning with training iterations. We can see that Figure~\ref{fig:b:cifar100_coef_pruning} has much more channels which are completely pruned away compared to Figure~\ref{fig:a:cifar100_coef_pruning}, which is evident from the fact that the former has a much higher MAC reduction than the latter.

\end{document}